\documentclass[runningheads]{llncs}
\usepackage{amssymb}
\usepackage{graphicx}
\usepackage{url}
\usepackage{color}
\usepackage{bm}

\usepackage{amsmath}
\usepackage{enumitem}
\usepackage{multirow}
\usepackage{subfig}
\usepackage{tikz,forest}
\usepackage{booktabs,arydshln}
\usepackage{pdflscape}
\usepackage{array}
\usepackage{makecell}
\usepackage{caption}
\usepackage{comment}
\usepackage[makeroom]{cancel}
\usepackage{pifont}
\newcommand{\cmark}{\ding{51}}%
\newcommand{\xmark}{\ding{55}}%

\newcommand{\PreserveBackslash}[1]{\let\temp=\\#1\let\\=\temp}
\newcolumntype{C}[1]{>{\PreserveBackslash\centering}p{#1}}
\newcolumntype{R}[1]{>{\PreserveBackslash\raggedleft}p{#1}}
\newcolumntype{L}[1]{>{\PreserveBackslash\raggedright}p{#1}}

\usepackage{tikz}

\definecolor{darkblue}{rgb}{0.0, 0.0, 0.55}
\definecolor{darkcyan}{rgb}{0.0, 0.55, 0.55}

\newcommand\blfootnote[1]{%
  \begingroup
  \renewcommand\thefootnote{}\footnote{#1}%
  \addtocounter{footnote}{-1}%
  \endgroup
}

\begin{document}
%

\title{An Evaluation Study of Intrinsic Motivation Techniques applied to Reinforcement Learning over Hard Exploration Environments}

\titlerunning{A Study on Intrinsic Motivation Techniques over Hard RL Environments}
%
%
\author{Alain Andres\inst{1,2}\and Esther Villar-Rodriguez\inst{1}\and Javier Del Ser\inst{1,2}}

\authorrunning{A. Andres et al.}
%
\institute{TECNALIA, Basque Research \& Technology Alliance (BRTA), 48160 Derio, Spain \\
\and University of the Basque Country (UPV/EHU), 48013 Bilbao, Spain \\
\email{\{alain.andres, esther.villar, javier.delser\}@tecnalia.com}}

\maketitle              
%

\begin{abstract}
In the last few years, the research activity around reinforcement learning tasks formulated over environments with sparse rewards has been especially notable. Among the numerous approaches proposed to deal with these hard exploration problems, intrinsic motivation mechanisms are arguably among the most studied alternatives to date. Advances reported in this area over time have tackled the exploration issue by proposing new algorithmic ideas to generate alternative mechanisms to measure the novelty. However, most efforts in this direction have overlooked the influence of different design choices and parameter settings that have also been introduced to improve the effect of the generated intrinsic bonus, forgetting the application of those choices to other intrinsic motivation techniques that may also benefit of them. Furthermore, some of those intrinsic methods are applied with different base reinforcement algorithms (e.g. PPO, IMPALA) and neural network architectures, being hard to fairly compare the provided results and the actual progress provided by each solution. The goal of this work is to stress on this crucial matter in reinforcement learning over hard exploration environments, exposing the variability and susceptibility of avant-garde intrinsic motivation techniques to diverse design factors. Ultimately, our experiments herein reported underscore the importance of a careful selection of these design aspects coupled with the exploration requirements of the environment and the task in question under the same setup, so that fair comparisons can be guaranteed.\blfootnote{Work published at the 6th International Cross Domain Conference for
Machine Learning \& Knowledge Extraction (CD-MAKE), 2022.}

\end{abstract}

%
%
%

\section{Introduction}
Over decades, Reinforcement Learning (RL) has been widely acknowledged as a rich and ever-growing research area within Artificial Intelligence aimed to efficiently deal with complex tasks \cite{silver2017mastering,baker2019emergent}. One of the key components of the success of RL algorithms is to define a suitable reward function that reflects the objective of the task at hand. However, the design of appropriate reward functions is often difficult -- even unfeasible -- depending on the peculiarities of the environment and task to be optimized. In this context, the study of environments with so-called \emph{sparse rewards} has gained attention in the last few years. In such scenarios,  the RL agent is just positively rewarded when accomplishing the goal, which is representative of manifold problems that arise from real-world applications \cite{holzinger2019introduction}. Nevertheless, such \emph{hard exploration} RL problems are more complex to address due to sparse informative feedback delivered by the environment, requiring effective means to balance between exploration and exploitation during the agent's learning process.

The aforementioned challenge can be overcome through Intrinsic Motivation (IM, \cite{aubret2019survey}), Imitation Learning \cite{ho2016generative} and Inverse Reinforcement Learning \cite{finn2016guided}, among other strategies. This work gravitates around the first (IM), which is used to encourage the agent to explore the environment by its inherent satisfaction of curiosity \cite{grigorescu2020curiosity}. In practice, the concept of \emph{curiosity} is translated to the RL domain in the form of an intrinsic bonus $r_i$, which is combined with the extrinsic reward provided by the environment $r = r_e + \beta r_i$ through a weighting factor $\beta$. Besides proposing different ways to generate the intrinsic reward $r_i$, current state-of-the-art algorithms (e.g., RIDE \cite{raileanu2020ride}, NGU \cite{badia2020never}, AGAC \cite{flet2021adversarially}) also apply novel methods to weight and scale such rewards, being those applicable to prior approaches such as Intrinsic Curiosity Module (ICM, \cite{pathak2017curiosity}) and Random Network Distillation (RND, \cite{burda2018exploration}). Unfortunately, when different IM-based schemes are compared to each other, those reward scaling techniques are not always in use, making it unclear whether the identified performance gaps are due to the exploration methods themselves or must be attributed to other design choices (i.e., the variation of intrinsic coefficient weights or the architecture of the models inside the RL agent; see Table \ref{tab:different_IM_methods}).
\begin{table}[h!]
    \centering
    \caption{Classification of multiple IM methods based on different design choices. We provide the parameters and modifications with which those approaches have been evaluated in MiniGrid \cite{gym_minigrid} benchmark except for NGU (Atari).} 
    \resizebox{\columnwidth}{!}
    {\begin{tabular}
     {L{2cm}C{2cm}C{2.2cm}C{2.2cm}C{4cm}}
     \toprule
     & \makecell[cb]{\texttt{RL-algorithm}} & \makecell[cb]{Vary $\beta_i$} & \makecell[cb]{Scale $r_i$} & \makecell[cb]{ANN architecture} \\
     \midrule
     ICM \cite{pathak2017curiosity} & IMPALA & \xmark & \xmark & Shared AC  [3CNN,256LSTM,FC]  \\
     RND \cite{burda2018exploration} & IMPALA & \xmark & \xmark & Shared AC, [3CNN,256LSTM,FC]  \\
     RIDE \cite{raileanu2020ride} & IMPALA & \xmark & \cmark & Shared AC, [3CNN,256LSTM,FC]  \\
     BeBold \cite{zhang2021noveld} & IMPALA & \xmark & \cmark & Shared AC, [3CNN,256LSTM,FC]  \\
     DoWhaM \cite{seurin2021don} & IMPALA & \xmark & \cmark & Shared AC, [3CNN,1024LSTM,1024FC]  \\
     AGAC \cite{flet2021adversarially} & PPO & \xmark & \cmark & Independent AC, [3CNN,512FC] \\
     D\&E \cite{jing2021divide} & PPO & \cmark & \cmark & Independent AC, [3CNN,512FC] \\
     RAPID \cite{zha2021rank} & PPO & \xmark & \xmark & Independent AC, [2FC64] \\
     \midrule
     NGU \cite{badia2020never} & R2D2 & \cmark & \cmark & Single Q(s,a,$\beta$), [4CNN,512LSTM,512FC]  \\
     \bottomrule
    \end{tabular}}
 \vspace{-3mm}
 \label{tab:different_IM_methods}
\end{table}

Analogously to what is claimed in other performance evaluation works reported recently in \cite{andrychowicz2020onpolicymatters,andrychowicz2020onpolicACymatters}, a fundamental matter in this research area is to discriminate which design criteria impact most on the performance of the RL agent. This is specially relevant in hard exploration environments, since it is known that under such circumstances, the proficiency of the agent is very sensitive regarding the configuration of its compounding modules. For this reason, this manuscript aims to fairly evaluate IM-based solutions present in the state of the art trying to decouple the solver approach from additional weighting and scaling techniques. Under this rationale, this work also incorporates the naive version of IM modules to study their benefit and ascertain the actual advantage of the algorithmic proposal when generating intrinsic rewards. Furthermore, the impact of having different neural network architectures in actor-critic agents and IM modules poses another question that lacks an informed answer in the current literature.

To sum up, this paper investigates the quantitative impact of different design choices when implementing IM-based techniques to understand their relevance when used in agents deployed over sparse reward scenarios. Hence, our contributions are three-fold: (1) we adopt curiosity mechanisms with different implementation choices that impact on how the intrinsic rewards are processed, (2) we conduct a study with multiple current state-of-the-art intrinsic motivation techniques where we compare them fairly in order to evaluate the improvement of generating rewards with different approaches; and (3) we break down experiments, results and conclusions in the interest of providing the reader with independent performance analysis of the set of modules and parameterizations.


The rest of the manuscript is structured as follows: Section \ref{sec:related_work} overviews works related to intrinsic motivation in RL, while Section \ref{sec:study_design} details the factors and the choices to be taken into account when resorting to IM techniques. Next, Section \ref{sec:experimental_setup} presents the experimental setup designed to achieve empirical evidence. Section \ref{sec:results} discusses the obtained results. Finally, Section \ref{sec:conclusion} concludes the paper and outlines future research to be developed from this research on.

\section{Related Work and Contribution} \label{sec:related_work}

Before proceeding with the details of this work, we briefly review insights coming from recent research about intrinsic motivation mechanisms to deal with sparse rewards. In the absence of a dense reward function and/or when having hard exploration problems, intrinsic motivation mechanisms have turned up as an effective workaround to overcome poor exploration behaviour. These techniques generate artificial intrinsic rewards based on the novelty of a state\footnote{Depending on the task under consideration, the novelty
can be associated to the very last performed action and/or the next state visited by the agent in the trajectory.}, which relates to how curious an agent will be when arriving to that state. The less novel a state is, the less curious the agent should be \cite{aubret2019survey}. In this context, several approaches have been proposed up to now to generate such exploration bonuses.

One mechanism to generate the aforementioned intrinsic rewards is by adopting a visitation count strategy, also referred to as \textit{count-based} methods. In this case, intrinsic rewards are assumed to be inversely proportional to the number of counts $N(s)$ that a given state $s$ has been visited, e.g. $r_{i}^{counts}=1/\sqrt{N(s)}$. This is a simple, yet effective, solution to quantify the degree to which a state is \emph{unknown} for the agent. However, counts are only applicable when dealing with discrete state spaces. Contrarily, when having more complex domains with continuous state spaces, density models \cite{bellemare2016unifying}, hash functions \cite{tang2017exploration} and also successor features \cite{machado2020count} can be applied to extend the concept of counts.


An alternative strategy to produce intrinsic motivation rewards is the use of \textit{prediction-error} methods which, as their name suggests, generate intrinsic rewards based on the error when predicting the consequence of an agent's action in the environment. The aforementioned Intrinsic Curiosity Module (ICM) proposed in \cite{pathak2017curiosity} belongs to this family of strategies, and operates by learning a state representation that just models the elements that the agent can control and those elements that can affect him. For this purpose, the intrinsic reward is generated based on the prediction error of the next state in a learned latent space:
\begin{equation}\label{eq:icm_rew}
    r_{i}^{ICM} = || \widehat{\phi}(s_{t+1}) - \phi(s_{t+1}) ||_2,
\end{equation}
where $\phi(\cdot)$ denotes the learned latent space mapping; $\widehat{\phi}(s_{t+1})$ is an estimation taking into account $\phi(s_t)$ and the actual action $a_t$; $s_t$ is the state visited at time $t$; and $||\cdot||_2$ stands for the $L_2$ (Euclidean) norm. 

Another approach is the use of RND introduced in \cite{burda2018exploration}. Under this strategy, two identical networks are randomly initialized, where one of the networks takes the role of predictor $\widehat{\phi}$ aiming to mimic the output of the other network -- namely, the target $\phi(\cdot)$, whose parameters are fixed after initialization. The reward is generated as an MSE loss between the outputs of both networks:
\begin{equation}\label{eq:rnd_rew}
    r_{i}^{RND} = || \widehat{\phi}(s_{t+1}) - \phi(s_{t+1}) ||^2.
\end{equation}

Built upon the idea of ICM, a recent work \cite{raileanu2020ride} introduced RIDE to use the same mechanism to learn the state embeddings, but they differ on how exploration bonuses are generated. In RIDE, this bonus is given by the difference between two consecutive states in their latent space:
\begin{equation}\label{eq:ride_rew}
r_{i}^{RIDE} = || \phi(s_{t+1}) - \phi(s_t) ||_2.
\end{equation}

With this change, RIDE encourages the agent to perform actions that have an impact on the environment. Moreover, by combining \textit{experiment-} and \textit{episode-level} \cite{pislar2021should} exploration to avoid the agent going back and forth between a sequence of states, the reward is discounted by the episodic state visitation counts:
\begin{equation}\label{eq:ride_ep_rew}
    r_{i}^{RIDE} = \frac{|| \phi(s_{t+1}) - \phi(s_t) ||_2}{\sqrt{N_{ep}(s_{t+1})}},
\end{equation}
where $N_{ep}(s_{t+1})$ denotes the episodic count of visits of state $s_{t+1}$. Following this idea of combining two levels of exploration (\textit{experiment-} and \textit{episode-}), the Never-Give-Up (NGU) approach in \cite{badia2020never} employs different intrinsic weights set in several parallel agents feeding the same network, which parameterizes each agent by making the neural network subject to the intrinsic coefficient used by each of them. A more aggressive strategy is BeBold/NoveID \cite{zhang2021noveld}, which goes beyond the boundaries of explored regions which only rewards (intrinsically) the first time the agent visits a given state in an episode. The Fast and Slow intrinsic curiosity in \cite{bougie2021fast} combines local and global exploration by generating two different intrinsic rewards, depending on the quality of the reconstruction of two contexts built from the same state. Furthermore, exploration can be enhanced by adversarially forcing an agent to solve a task in different ways \cite{flet2021adversarially,campero2020learning}.

Rather than proposing a new intrinsic generation module, the present work offers a study combining different design choices made in recent solutions and fairly compare them under equal experimental conditions. This being said, other benchmarks/studies have been done in recent times: to begin with, \cite{taiga2021bonus} evaluates the performance of different exploration bonuses (pseudo-counts, ICM, RND and noisy networks) in the whole Atari 2600 suite with Rainbow \cite{hessel2018rainbow}. By contrast, \cite{burda2018large} carried out a large-scale study based exclusively on prediction error bonuses (ICM) over 54 environments, where they investigated the efficacy of using different feature learning methods with Proximal Policy Optimization (PPO, \cite{schulman2017proximal}). Our work also connects with \cite{andrychowicz2020onpolicymatters,andrychowicz2020onpolicACymatters,orsini2021matters}, a series of evaluation studies aimed to understand what choices among high- and low-level algorithmic options affect the learning process: as such, the studies in \cite{andrychowicz2020onpolicymatters,andrychowicz2020onpolicACymatters} focus on on-policy deep actor-critic methods (examining different policy losses, architectures and advantage estimators), whereas \cite{orsini2021matters} addresses Adversarial Imitation Learning related decisions (multiple reward functions and observation normalization methods).
 
\paragraph{Contribution:} To the best of our knowledge, there is no prior work that exhaustively evaluates different choices for the implementation of intrinsic motivation strategies. Our study takes a step further by analyzing different weight and scale strategies for the combination of intrinsic and extrinsic rewards, as well as the impact of adopting different neural networks architectures and dimensions. The design choices here evaluated are applicable to any intrinsic curiosity generation module, so that conclusions about which ones are the most suitable given a task and an environment with sparse rewards can be drawn.

\section{Methodology of the Study}\label{sec:study_design}

After reviewing different solutions proposed in the literature to cope with hard exploration issues with IM techniques, we now proceed by describing the methodology adopted in this study to gauge the advantages and drawbacks of design choices that are present in some of them, giving an informed hint of their utility when extrapolated to the rest of IM solutions. The methodology is driven by the pursuit of responses to three research questions (RQ):
\begin{itemize}[leftmargin=*]
    \item RQ1: Does the use of a static, parametric or adaptive decaying intrinsic coefficient weight $\beta$ affect the agent's training process?
    \item RQ2: Which is the impact of using episodic counts to scale the intrinsic bonus? Is it better to use episodic counts than to just consider the first time a given state is visited by the agent?
    \item RQ3: Is the choice of the neural network architecture crucial for the agent's performance and learning efficiency?
\end{itemize}

\noindent Departing from these questions, the following methodology has been devised:

\subsection{RQ1: Varying the Intrinsic Reward Coefficient $\beta$} \label{subsec:int_rew_weight}

In general, it is not advisable to combine raw extrinsic and intrinsic reward signals directly due to their potentially diverging value scales. Moreover, even if taking values from comparable ranges, the agent could need to grant more importance to exploration than to exploitation at specific periods. In fact, in sparse rewards settings, the explorer role of the agent must be strengthen and enlarged in comparison to the exploitative behaviour to guide the agent by an artificial bonus in the absence of knowledge about the target task. This balance between exploration and exploitation is usually controlled by the intrinsic reward coefficient $\beta$, whose value is often tuned manually depending on the environment and task to be accomplished. A priory, this value might be fixed and kept unaltered, or dynamically updated, as is further explained in what follows:
\paragraph{\textbf{Static} $\bm{\beta}$:} commonly, the $\beta$ coefficient is stationary along the whole training. In such cases,  we refer to this fixed and default value as $\beta_s$. On this basis, diverse fixed intrinsic coefficient values can be used to learn a family of policies with different exploration-exploitation balances, so as to concentrate on maximizing the extrinsic reward (a policy with $\beta=0$) while maintaining a degree of exploration (rest of policies with $\beta > 0$) \cite{badia2020never}. Contrarily to the rest of approaches, when using multiple (fixed) intrinsic coefficients training more than one agent is required. 

\begin{figure}[ht]
    \centering
    \includegraphics[width=0.7\columnwidth]{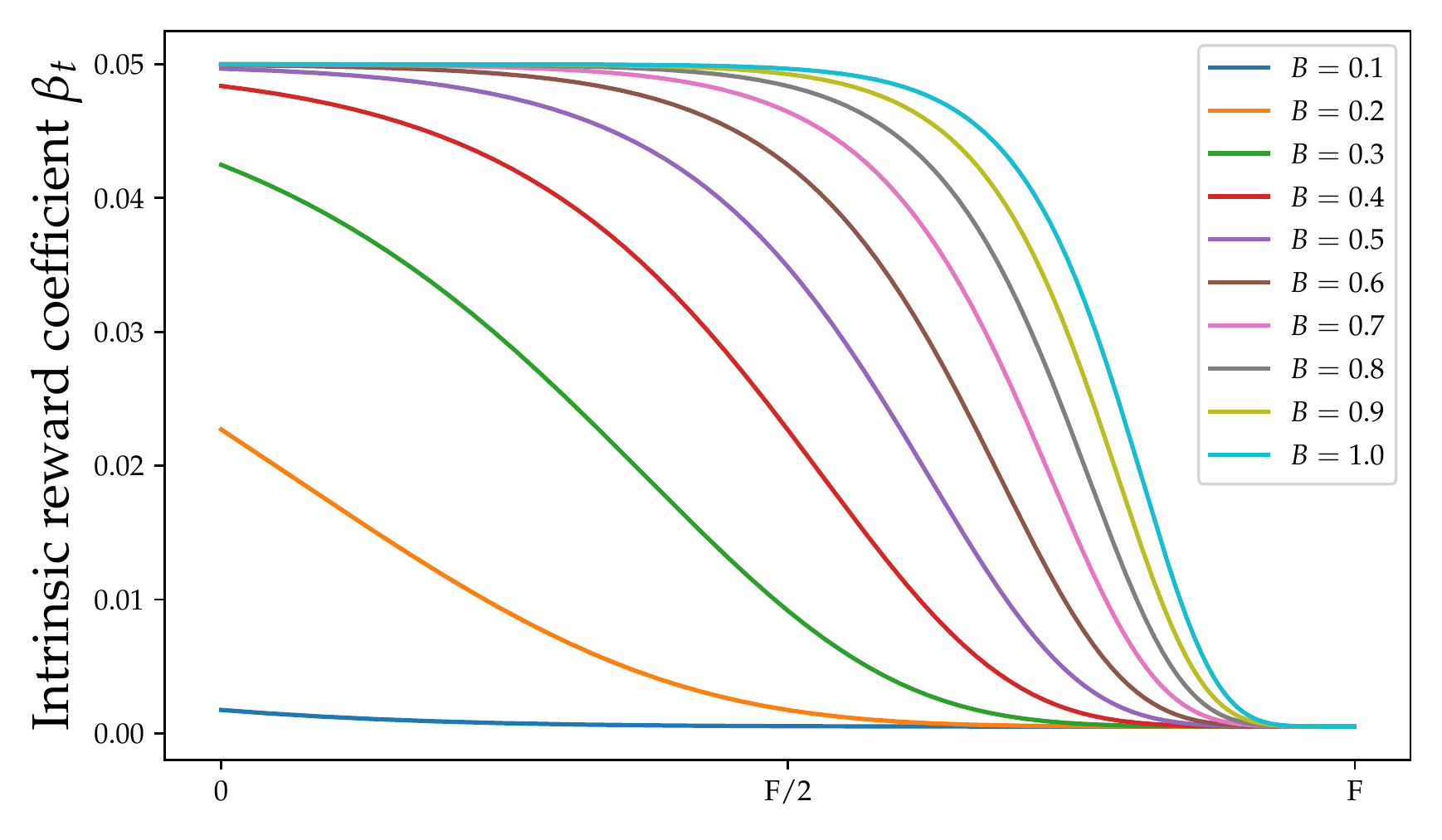}\vspace{-3mm}
    \caption{Example of the parametric decay evolution of $\beta_t$ for multiple values of the smoothness control parameter $B$, with $K=0.05, A=0.0005$ and $F=2e7$.} 
    \vspace{-3mm}
    \label{fig:parametric_decay_curve}
\end{figure}

\paragraph{\textbf{Dynamic} $\bm{\beta}$:} to focus on the extrinsic signals provided by the environment, it is interesting to modulate the weight given to the intrinsic rewards generated by the agent in a dynamic fashion. Without loss of generality, in our work we consider two different options: parametric decay and adaptive decay. For the \emph{parametric} decay, the value of $\beta$ decreases by following a modified sigmoid function, which parametrically controls the smoothness of the decay:
\begin{equation} \label{eq:beta_dynamic_richard_curve}
    \beta_t = A + \frac{K-A}{\left(1 + \exp\left(-16B\left(1-\frac{t}{F}\right)\right)\right)^{20}}
\end{equation}
where $K$ is a value proven to deliver a good performance and well balanced trade-off between exploration and exploitation (e.g., the fixed value $\beta_s$ that one could select under a fixed $\beta$ strategy); $A$ is the final value of $\beta$, which can be defined from $K$ (e.g. $A=K/100$) to reflect that at the end of the learning process, the agent should receive hardly any intrinsic signal bonus; and $F$ denotes the number of frames (= sample, steps) we expect the whole train to have. Moreover, $B$ permits to control the \emph{smoothness} of the progression of $\beta$ throughout the training (Figure \ref{fig:parametric_decay_curve})\footnote{Note that this parametric decay can also be used to sample different $\beta$ values for each policy learned by means of the approach with multiple static intrinsic coefficients $\beta$, by defining $F$ as the number of agents.}.

In turn, we can vary the intrinsic coefficient by adopting an \texttt{adaptive decay strategy}. Motivated by \cite{jing2021divide} for concurrent environments, we propose to calculate a decay factor $d_i^\tau$\footnote{Rollout is denoted as $\tau$, whereas the \textit{i-th} rollout is denoted as $\tau_i$.Note that $i \neq t$, as a rollout comprehends more than one time step ($t$); \textit{i} is used to refer to rollouts.} based on the ratio between the agent's intrinsic return at the current rollout, $G_i^\tau$, and the averaged historical intrinsic return values in past rollouts $H^\tau$: 
\begin{equation} \label{eq:adaptive}
    \beta_i^\tau = \beta_s d_i^\tau =\beta_s \min \left[\frac{G_i^\tau}{H^\tau}, 1 \right]  =\beta_s \min \left[\frac{G_i^\tau} {\frac{1}{K}\sum_{k=0}^{K=i} G_k^\tau},1\right],
\end{equation}
where $K$ is the total number of rollouts the agent collected from the beginning of the training process up to the present rollout $i$. Consequently, under this rationale the agent is discouraged from exploring those trajectories that are more familiar than the average and means less novelty. Furthermore, the intrinsic return during the training may vary due to the non-stationary nature of the intrinsic reward generation process. Thereby, to stabilize the training, instead of leveraging the whole historical data, we also propose the use of a moving average with a sliding window, $H_\omega^\tau$, which strictly considers just the latest returns ($\omega$) and avoids the case of discouraging the exploration due to large initial intrinsic returns that may well bias the decay factor calculation.

\subsection{RQ2: Episodic State Counts versus First-Visit Scaling} \label{subsec:int_rew_scale}
As defined in \cite{pislar2021should}, there are different periods in which the exploration mode can be carried out: \textit{step-level}, \textit{experiment-level}, \textit{episode-level}, or \textit{intra-episodic}. Over the years the use of  \textit{step-level} exploration (i.e. $\epsilon$-greedy) has proven to yield good results in a diversity of simple RL environments. However, advances in learning algorithms have paved the way towards RL problems of higher complexity, in which the exploration is one of the critical parts to be addressed. As has been already argued in the introduction, hard exploration problems can be tackled by letting the agent explore the environment by its inherent satisfaction (intrinsic motivation) rather than being guided by environment provided extrinsic feedback signals. Nevertheless, intrinsic motivation techniques are prone to a quick vanishing of the rewards over the course of the training, reducing attractiveness as the training evolves. This condition is exacerbated when facing long-time horizon problems \cite{bougie2021fast}. Actually, by analyzing the rewards obtained during a concrete episode, few differences in terms of novelty are appreciated between similar/close states, even if one has been already visited and the other remains unexplored. This is due to the persistence of curiosity-related information from past episodes (\textit{experiment-level}), which is propagated forward during the agent's training leaving little novelty difference between similar (even identical) states inside the scope of the same episode. Additionally, in environments where state transitions are reversible, using intrinsic rewards to guide the exploration can lead into an agent bouncing back and forth between sequences of states that are more novel than others in the same episode \cite{raileanu2020ride,zhang2021noveld}. 

As a solution to this issue, recent studies \cite{raileanu2020ride,badia2020never,zhang2021noveld,bougie2021fast,seurin2021don} have combined two degrees of novelty rather than just one: local (\textit{episode-level}) and global (\textit{experiment-level}). More concretely, \cite{raileanu2020ride} employed an episodic visitation count term to encourage the agent to visit as many different states as possible within an episode. Similarly, \cite{zhang2021noveld} incorporated a more aggressive variation that rewards the agent only when it visits a given state for the first time within an episode. However, approaches at the forefront of the state of the art (i.e., ICM, RND) do not implement this idea to scale their rewards. In this context, it is unclear whether new proposed IM modules outperform previous approaches due to state-count regularization or to conceptually new algorithmic schemes. If state-counts regularization contributed to improve the performance, already proposed IM schemes that do not implement it and also future IM methods could adopt this strategy to meliorate their designs. Hence, our experimentation studies the above mentioned two episodic cases.

\subsection{RQ3: Sensitiveness to the Neural Network Architectures} \label{subsec:architecures}

In the literature related to RL, plenty of network architecture proposals have been used to solve any given problem. As an example, the work in \cite{zha2021rank} simplified the architectures previously proposed in \cite{raileanu2020ride}, yet achieving similar results\footnote{We note that the choice of the neural network architecture is not just for the actor-critic modules, but also for IM approaches that hinge on neural computation.}. However, they rely on different base RL algorithms (PPO \cite{schulman2017proximal} and IMPALA \cite{espeholt2018impala}, respectively), thereby hindering a fair comparison, a proper interpretability and attribution of the reported performance results (check Table \ref{tab:different_IM_methods}).

To avoid this issue, our specific experimentation evaluates the effect of the network architecture on the performance of the RL agent by considering a fixed RL algorithm and IM module, and by assessing several network configurations. By reporting the dimensions and characteristics of different neural network architectures and the performance of RL agents using them, we can gain intuition about the performance improvement (degradation) incurred when increasing (decreasing) the complexity of the neural architectures in use. Our experiments also measure the required amount of time when using those architectures, so that latency implications can be examined. This third research question is also aligned with practical concerns arising when deciding on which implementation is more suitable for a real-world deployment, specially in resource-constrained scenarios (e.g. embedded robotic devices).  

\section{Experimental Setup} \label{sec:experimental_setup}

We answer RQ1, RQ2 and RQ3 over procedurally generated RL tasks from the Minimalistic Gridworld Environment (MiniGrid \cite{gym_minigrid}). This framework allows creating RL tasks of varied levels of difficulty, does not strictly make use of images as observations, and most importantly, runs fast, thereby easing the implementation of massive RL benchmarks. 

\subsection{Environments}
\begin{figure}
    \centering
    \vspace{-3mm}
        \subfloat[\texttt{MN7S8}]{      \includegraphics[width=0.33\columnwidth]{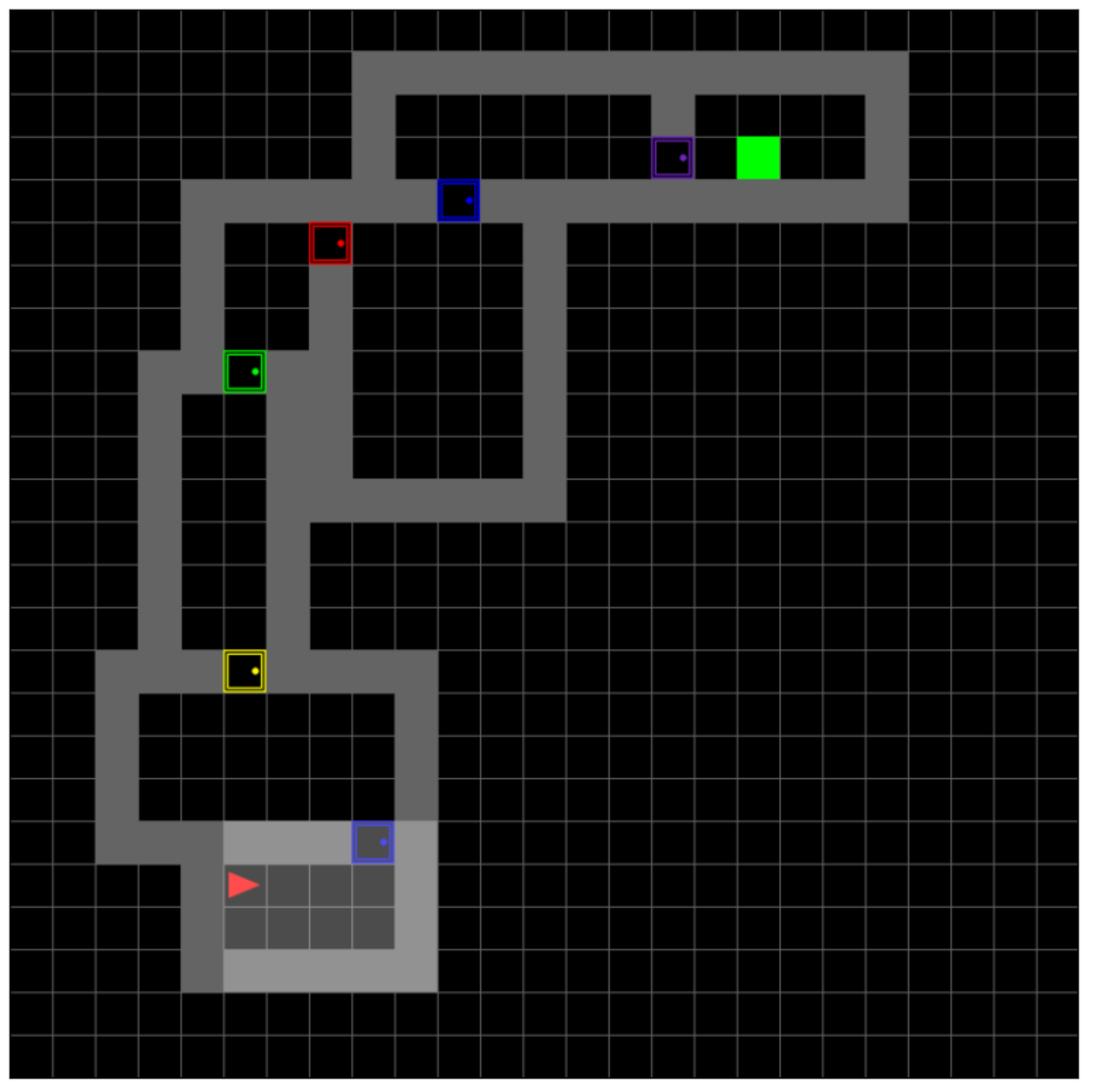}}\:
        \subfloat[\texttt{KS3R3}]{\includegraphics[width=0.33\columnwidth]{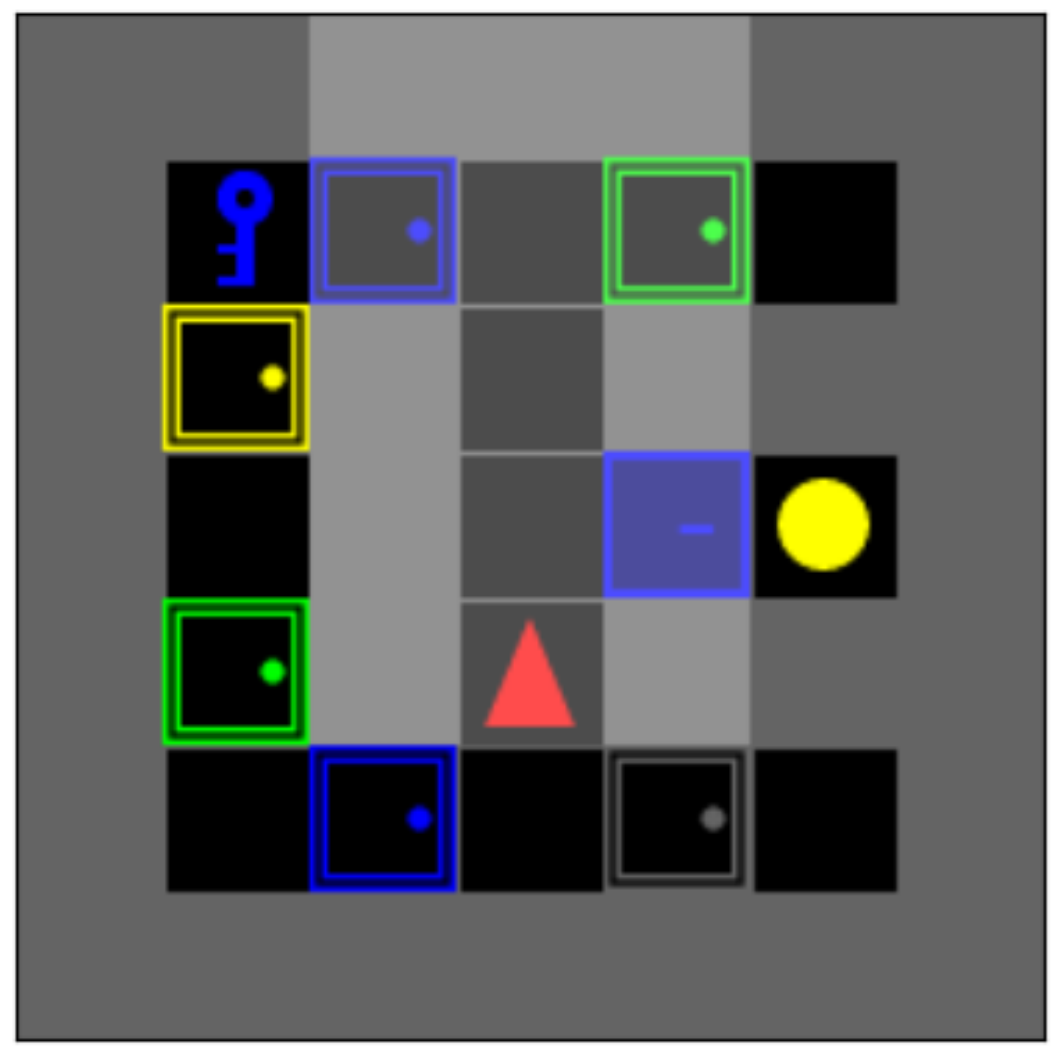}}\:
        \subfloat[\texttt{O2Dlh}]{
        \includegraphics[width=0.123\columnwidth]{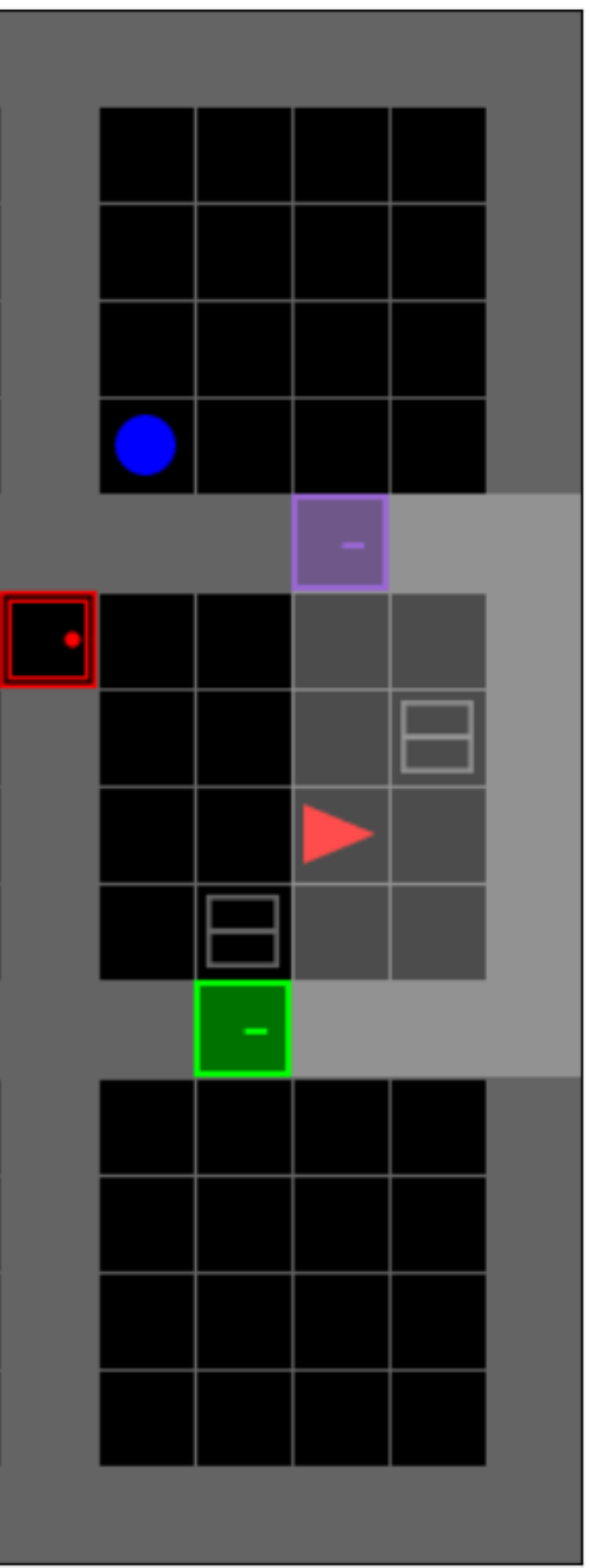}
        \:\includegraphics[width=0.122\columnwidth]{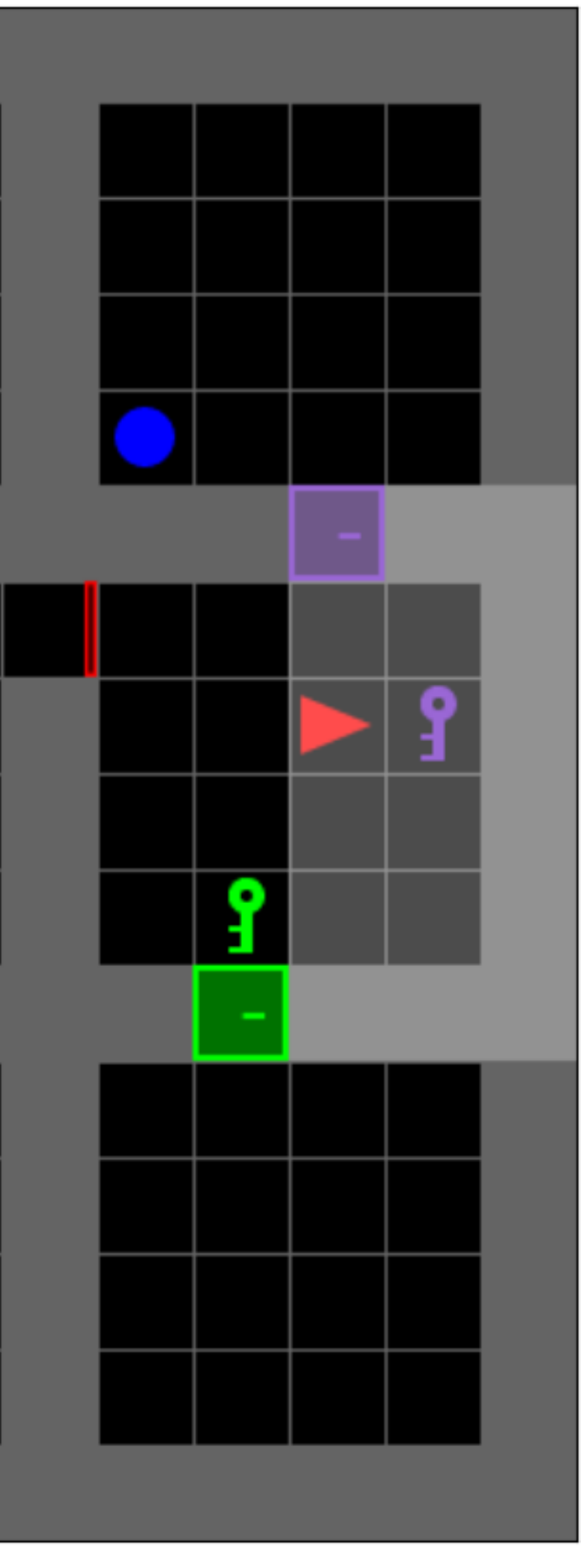}}
    \caption{Examples of MiniGrid scenarios (a) \texttt{MN7S8}: the agent has to open multiple doors to reach the distant goal (green square); (b) \texttt{KS3R3}: the agent has to first collect the blue key in order to open the door of the room leading to the yellow ball that must be picked (c) \texttt{O2Dlh}: the agent has to discover keys hidden below some boxes, take the proper key and open the door to the blue ball (target).}
    \vspace{-3mm}
    \label{fig:minigrid_envs}
\end{figure}

To design a representative benchmark for the study, among all the possible RL environments that can be selected/generated in MiniGrid, we consider 1) those labeled as \texttt{MultiRoomNXSY} (shortened as \texttt{MNXSY}, with \texttt{X} denoting the number of rooms and \texttt{Y} their size), 2) \texttt{KeyCorridorS3R3} (\texttt{KS3R3}); and 3) \texttt{ObstructedMaze2Dlh} (\texttt{O2Dlh}). These scenarios belong to hard exploration tasks (i.e., rewards are sparse), in which the agent fails to complete the task without the help of any IM mechanism. Refer to Figure \ref{fig:minigrid_envs} for further information about each scenario and its associated goal. 

By default, observations in these tasks are essentially egocentric and partially observable views of the environment, where a $7\times 7$ tile set in the direction that the agent is facing composes the observation. Concretely, an observation is featured by a 7x7x3 matrix, being the 3 features of the last dimension information of interest such as type, colour and status of the object (e.g., doors, keys, balls, or walls) placed in the specific tile. Notice that the agent is incapable to see through walls or doors. 7 basic actions are available to solve all scenarios: \texttt{turn left}, \texttt{turn right}, \texttt{move forward}, \texttt{pick up} (an object, for instance keys or balls), \texttt{drop} the object (if carried), \texttt{toggle} (open doors, interact with objects) and \texttt{done}. Nevertheless, some of these actions are only useful at specific locations, whereas others become useless for certain tasks (for instance, \texttt{pick}/\texttt{drop} and \texttt{done} in \texttt{MNXSY} environments).

Not all the environments require the same amount of steps to be solved. Thus, in \texttt{MNXSY} environments a maximum number of $20\cdot \texttt{X}$ steps is set to make it dependent on the number of rooms. Consequently, the three considered environments that fall within this set (\texttt{MN7S4}, \texttt{MN7S8} and \texttt{MN10S4}) are assumed to take at most 140, 140 and 200 steps, respectively. For \texttt{KS3R3} $270$ and for \texttt{O2Dlh} $576$ steps are set as maximum. The rewards are valued according to the number of steps taken. The optimal average extrinsic returns that the agent can achieve are $0.77$ (\texttt{MN7S4}), $0.76$ (\texttt{MN10S4}), $0.65$ (\texttt{MN7S8}), $0.9$ (\texttt{KS3R3}), and $0.95$ (\texttt{O2Dlh}). Actually, since they are procedurally generated environments, each scenario's final reward can slightly change due to the variance on the minimum required steps. In our case, we get these values by taking the median value of an optimal policy (equal to other previous reported optimal results \cite{zhang2021noveld}). Moreover, we also refer as suboptimal behavior to those policies that managed to obtain at least a 95\% of the optimal score. In terms of complexity, \texttt{MN7S4} and \texttt{MN10S4} are the easiest ones to solve, followed by \texttt{MN7S8} and \texttt{KS3R3} which are harder. Finally, \texttt{O2Dlh} is the most difficult task in our benchmark.

\subsection{Baselines and Hyperparameters}
All our experiments will employ PPO \cite{schulman2017proximal} as the main RL algorithm. On top of it, we will use state-of-the-art IM techniques in order to obtain intrinsic rewards to augment the exploration efficiency, in which a naive PPO model fails \cite{zha2021rank}: COUNTS\footnote{In this case, we take advantage of the 2D grid (discrete state space) and map each state directly to a dictionary when using COUNTS. Nevertheless, when facing more complex state spaces pseudo-counts \cite{bellemare2016unifying} can be applied as an alternative as in \cite{taiga2021bonus}.}, RND \cite{burda2018exploration}, ICM \cite{pathak2017curiosity} and RIDE \cite{raileanu2020ride}. For PPO we use a discount factor $\gamma$ equal to $0.99$, a clipping factor $\epsilon=0.2$, $4$ epochs per train step and $\lambda =0.95$ for GAE \cite{schulman2015high}. We use 16 parallel environments to gather rollouts of size 128. Hence, we set a total horizon of $2,048$ steps between updates. Moreover, a batch size equal to 256 is considered. Unless otherwise specified, the following values - selected from an off-line grid search procedure over \texttt{MN7S4} - will be used to configure the intrinsic coefficient and entropy: $\beta = 0.05$ and $\varepsilon = 0.0005$ for RND, ICM and RIDE; $\beta = 0.005$ and $\varepsilon = 0.0005$ for COUNTS. In what refers to the dynamic update of $\beta$, we select $B=0.5$ in Expression \eqref{eq:beta_dynamic_richard_curve} as it represents a balanced trade-off for the agent to explore in the early stages of the training process, evolving towards a behavior mainly driven by extrinsic signals.

\subsection{Network Architectures}\label{subsec:network_arch}

Finally, experiments around RQ3 are performed with two different neural network architectural designs, which differ in terms of the type of neural layers (and design) and their number of trainable parameters. Following Figure \ref{fig:archs}, on one hand a \emph{lightweight} neural architecture as in RAPID\cite{zha2021rank} is considered, in which both the actor and the critic are made of 2FC with 64 neurons each. This dual FC-64 architecture also applies to the embedding networks required for RND, ICM and RIDE. Additionally, we include a more sophisticated neural design based on what is proposed in RIDE \cite{raileanu2020ride}, where both the actor and critic are combined into a two-headed (one for the policy, the other for the critic) shared network with 3 convolutional neural layers (32 $3\times 3$ filters, stride equal to 2, and padding 1) and a FC-256 layer\footnote{Originally in \cite{raileanu2020ride} they used a LSTM of 256 units instead of a FC-256. However, even if they claim the use of such architecture in the paper, in their github implementation they seem to use larger networks \url{https://github.com/facebookresearch/impact-driven-exploration}, reason why in Table \ref{tab:different_IM_methods} we do not specify the FC units.}. This last architecture will be deemed the \emph{default} architecture to endow the agent with more learning capabilities and to ensure that it is not limited by a restricted network.
\begin{figure}[h!]
    \centering
\includegraphics[width=0.8\columnwidth]{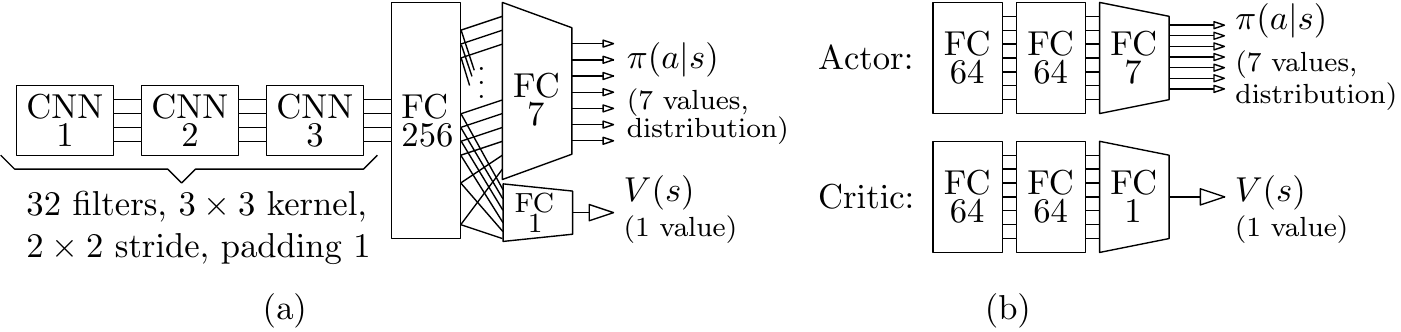}
\caption{(a) Sophisticated/default and (b) lightweight network architectures.}
\label{fig:archs}
\end{figure}

\section{Results and Analysis} \label{sec:results}

This section is devoted to present experiment results and answer the research questions posed
in Section \ref{sec:study_design}. Scripts and results have been made available in a public GitHub repository (\url{https://github.com/aklein1995/intrinsic_motivation_techniques_study}) to foster reproducibility and follow-up studies. For all the experiments described in this section we provide the mean and standard deviation of the average return computed over the past 100 episodes, performing 3 different runs (each with a different seed) to account for the statistical variability of the results.

\subsection*{RQ1: Does the use of a static, parametric or adaptive decaying intrinsic coefficient weight $\bm{\beta}$ affect the agent's training process?}

Our first set of results compares the multiple weighting strategies introduced in Section \ref{subsec:int_rew_weight}, which differently tune the importance granted to the intrinsic rewards with respect to extrinsic signals coming from the environment. 

The results are shown in Table \ref{table:rq1}, where it is straightforward to note that RIDE outperforms COUNTS and RND. At this point we remind the reader that RIDE is configured with episodic count scaling, in accordance with the final solution proposed in \cite{raileanu2020ride}. Count-based generated rewards seem to be the best solution when facing easy exploration scenarios (\texttt{MN7S4} and \texttt{MN10S4}), but its performance degrades when facing scenarios that require more sophisticated exploration strategies. A similar pattern can be observed when analyzing the results of RND, which is unable to solve \texttt{MN7S8} and \texttt{O2Dlh} with any kind of weighting strategy. Contrarily, RIDE manages to solve all the tasks by its basic implementation, although it obtains better results when using more sophisticated weighting exploration strategies.

\begin{table}[t]
    \centering
    \caption{Results of different IM strategies over several MiniGrid scenarios with static ($\_s$), multiple static ($\_ngu$) (as in NGU \cite{badia2020never}), a parametric ($\_pd$) or adaptive decay ($\_ad$) weight $\beta$ to modulate the importance of the intrinsic bonus in the computation of the reward. Cell values denote the training steps/frames ($1e6$ scale) at which the optimal average extrinsic return is achieved; in parentheses, steps at which $95\%$ of the optimal average extrinsic return is reached. Best results for every (IM strategy, scenario) combination are highlighted in bold.} 
    \resizebox{\columnwidth}{!}
    {\begin{tabular}{L{2.5cm}C{2cm}C{2cm}C{2.2cm}C{2.2cm}C{2.2cm}}
     \toprule
     & \makecell[cb]{\texttt{MN7S4}} & \makecell[cb]{\texttt{MN10S4}} & \makecell[cb]{\texttt{MN7S8}} & \makecell[cb]{\texttt{KS3R3}} & \makecell[cb]{\texttt{O2Dlh}} \\
     \midrule
     COUNTS$\_s$  & \textbf{0.93} (0.86) & 1.87 (1.78) & $>$ 30 & $>$ 30 & $>$ 50 \\
     COUNTS$\_ngu$ & 1.17 (1.11) & 2.67 (2.35) & $>$ 30 & $>$ 30 & $>$ 50 \\
     COUNTS$\_pd$ & 0.96 \textbf{(0.83)} & 2.27 (1.67) & $>$ 30 & \textbf{22.91 (22.49)} & $>$ 50 \\
     COUNTS$\_ad$ & 1.03 (0.92) & \textbf{1.81 (1.65)} & 24.23 (24.10) & $>$ 30 & $>$ 50 \\
     COUNTS$\_ad1000$ & 1.03 (0.92) & \textbf{1.81 (1.65)} & 23.63 (23.56) & $>$ 30 & $>$ 50\\
     \midrule
     RND$\_s$  & 3.83 (3.78) & 7.84 (7.79) & $>$ 30 & 10.83 (9.72) & $>$ 50 \\
     RND$\_ngu$ & 2.69 (2.62) & 5.78 (5.75) & $>$ 30 & 8.12 (7.50) & $>$ 50 \\
     RND$\_pd$ & 4.04 (3.94) & 6.02 (5.99) & $>$ 30 & 9.24 (8.07) & $>$ 50 \\
     RND$\_ad$ & \textbf{2.02 (1.39)} & \textbf{3.21 (2.65)} & $>$ 30 & \textbf{6.02 (5.43)} & $>$ 50 \\
     RND$\_ad1000$ & 3.62 (1.42) & 3.59 (3.50) & $>$ 30 & 7.47 (6.66) & $>$ 50 \\
     \midrule
     RIDE$\_s$  & \textbf{2.49} (1.82) & 2.27 (2.14) & 4.00 (3.68) & 6.63 (4.39) & 30.88 (25.87) \\
     RIDE$\_ngu$ & 3.85 (2.40) & 2.59 (1.26) & $>$ 30 & 7.18 (3.91) & 36.07 (29.96) \\
     RIDE$\_pd$ & 5.20 (2.14) & 5.01 (1.96) & \textbf{3.73 (3.49)} & 6.42 (3.87) & 29.27 \textbf{(20.84)} \\
     RIDE$\_ad$ & 2.89 \textbf{(0.91)} & \textbf{1.60} \textbf{(0.99)} & $>$ 30 & 5.93 (2.99) & \textbf{27.65} (20.91) \\
     RIDE$\_ad1000$ & 2.54 \textbf{(0.91)} & \textbf{1.60} \textbf{(0.99)} & 3.88 (3.70) & \textbf{4.70 (3.00)} & 28.00 (23.01) \\
     \bottomrule
    \end{tabular}}
    \vspace{-3mm}
 \label{table:rq1}
\end{table}

We now focus the discussion on gaps arising from the use of different weighting strategies. The static (default) weighting strategy (indicated with a suffix $\_s$ appended to each approach) is surpassed by any of the other proposed weighting approaches in the majority of the cases. When using multiple static values ($\_ngu$), the only approach that takes advantage of such strategy is RND, yielding worse results for both COUNTS and RIDE in all the cases. This might happen due the slow pace at which the intrinsic rewards values decay in RND in reference to the other strategies. Moreover, the error outputs higher amplitude values than those of RIDE, then being RND a better candidate to get benefit of applying the $\_ngu$ strategy by the use of agents with smaller intrinsic coefficient weights. On the other hand, the use of parametric decay ($\_pd$), which decreases the weight of the intrinsic reward as the train progresses to favor exploration, provides significant gains in almost all simulated scenarios. This approach is similar to $\_ngu$ although, instead of using multiple agents with different static intrinsic coefficients, it modulates a single value during the course of training. Hence, when employing $\_pd$ strategy, COUNTS is able to get a valid solution in \texttt{KS3R3}, RND improves all its scores and RIDE improves its behaviour in the most challenging scenarios \texttt{MN7S8, KS3R3 and O2Dlh}. Nevertheless, $\_ngu$ and $\_pd$ highly depend on the intrinsic coefficients given to each agent and the evolution of a single intrinsic coefficient during training, respectively. This strongly impacts on the agent's performance for a given scenario and dictates when those approaches might be better.

Finally, the use of adaptive decay ($\_ad)$ produces better results in COUNTS and RND when compared to the static case ($\_s$). For RIDE, however, this statement does not strictly hold true, as its performance degrades in \texttt{MN7S4} and \texttt{MN7S8} (the agent does not even solve the task in the latter). We hypothesize that this is because the initial intrinsic returns are too high and calculating the historical average intrinsic returns biases the decay factor calculation. As outlined in Section \ref{subsec:int_rew_scale}, a workaround to bypass this issue is to calculate returns with a moving average over a window of $\omega$ steps/rollouts. We hence include in the benchmark an adaptive decay with a window size of $\omega=1000$ rollouts ($\_ad1000$). With this modification, RIDE improves its behavior in all the complex scenarios. Nevertheless, $\_ad1000$ performs slightly worse than $\_ad$ in RND, but never worse than its static counterpart $\_s$. In general, $\_ad1000$ promotes higher intrinsic coefficient values than $\_ad$, as the calculated average return is better fit to the actual return values. This leads to a lower decay value and a higher intrinsic coefficient, forcing the agent to explore more intensely than with $\_ad$ (but less than with $\_s$).

\subsection*{RQ2: Which is the impact of using episodic counts to scale the intrinsic bonus? Is it better to use episodic counts than to just consider the first time a given state is visited by the agent?}

Answers to this second question can be inferred from the results of Table \ref{table:rq2}. A first glance at this table reveals that the use of episodic counts or first-time visitation strategies for scaling the generated intrinsic rewards leads to better results. In the most challenging environments (\texttt{MNS78}, \texttt{KS3R3} and \texttt{O2Dlh}), these differences are even wider, as they require a more intense and efficient exploration by the agent. In fact, when the training stage is extended to cope with the resolution of a more complex task, intrinsic rewards also decrease, inducing a lower explorative behaviour in the agent the more the train is lengthened. What is more, the agent is not encouraged to collect/visit as many different states as possible. Hence, in those scenarios the baseline implementation of intrinsic motivation ($\_noep$) may fail, but with these scaling strategies the problem is resolved (i.e. COUNTS and RND in \texttt{O2Dlh}). By contrast, in environments requiring less exploration (\texttt{MN7S4} and \texttt{MN10S4}), differences are narrower when using \textit{episode-level} exploration and may be counterproductive in some cases (i.e. COUNTS at \texttt{MN10S4} with $\_1st$). 
\begin{table}[ht!]
    \centering
    \vspace{-3mm}
    \caption{Comparison of different IM strategies when using no scaling ($\_noep$), episodic ($\_ep$) or first-time visit ($\_1st$) to scale the generated intrinsic reward and combine two types of exploration degrees. Interpretation as in Table \ref{table:rq1}.} 
    \resizebox{\columnwidth}{!}
    {\begin{tabular}{L{2.5cm}C{2cm}C{2cm}C{2.2cm}C{2.2cm}C{2.2cm}}
     \toprule
     & \makecell[cb]{\texttt{MN7S4}} & \makecell[cb]{\texttt{MN10S4}} & \makecell[cb]{\texttt{MN7S8}} & \makecell[cb]{\texttt{KS3R3}} & \makecell[cb]{\texttt{O2Dlh}} \\
     \midrule
     COUNTS$\_noep$  & 0.93 (0.86) & 1.87 (1.78) & $>$ 30 & $>$ 30 & $>$ 50 \\
     COUNTS$\_ep$    & \textbf{0.76} (0.56) & \textbf{1.55 (1.47)} & 2.77 (2.56) &  3.99 (2.00)& \textbf{33.17 (29.79)} \\
     COUNTS$\_1st$   & 0.85 \textbf{(0.48)} & $>$ 20 & \textbf{1.64 (1.42)} & \textbf{1.97 (1.19)} & 45.26 (37.29) \\
     \midrule
     RND$\_noep$  & 3.83 (3.78) & 7.84 (7.79) & $>$ 30 & 10.83 (9.72) & $>$ 50 \\
     RND$\_ep$    & 1.41 (0.96) & 1.72 (1.34)& 3.60 (3.30) & \textbf{4.31} (2.63) & \textbf{18.54} (14.07) \\
     RND$\_1st$   & \textbf{1.18 (0.59)} & \textbf{1.36 (0.78)} & \textbf{1.97 (1.72)} & 4.78 \textbf{(2.29)} & 21.19 \textbf{(9.88)} \\
     \midrule
     RIDE$\_noep$   & 4.71 (4.54) & 5.29 (5.20) & $>$ 30 & 11.44 (9.63) & 39.68 (35.15) \\
     RIDE$\_ep$     & \textbf{2.49} (1.82) & \textbf{2.27 (2.14)} & 4.00 (3.68) & 6.63 (4.39) & \textbf{30.88 (25.87)} \\
     RIDE$\_1st$    & 3.17 \textbf{(1.34)} & 3.27 (2.33) & \textbf{1.95 (1.83)} & \textbf{5.13 (2.26)} & 32.14 (28.03) \\
     \midrule
     ICM$\_noep$  & 2.67 (2.55) & $>$ 20 & $>$ 30 & 8.02 (6.75) & 34.04 (26.78) \\
     ICM$\_ep$    & 3.25 (1.26) & \textbf{1.68} (1.59) & $>$ 30 & 5.32 (3.14) & \textbf{19.05} (13.87) \\
     ICM$\_1st$   & \textbf{1.56 (0.87)} & 1.90 \textbf{(1.07)} & \textbf{2.11 (1.77)} & \textbf{4.72 (4.23)} & 20.74 \textbf{(10.09)} \\
     \bottomrule
    \end{tabular}}
  \vspace{-3mm}
 \label{table:rq2}
\end{table}

To better understand the superiority of RIDE over ICM \cite{raileanu2020ride}, we also evaluate the performance of both approaches under equal conditions, with ($\_ep$, $\_1st$) and without ($\_noep$) scaling strategies. In this way, we can examine the actual improvement between the two types of exploration bonus strategies. Surprisingly, ICM gives better results in almost all the cases for the analyzed scenarios, yet exhibiting a larger variance in several environments that lead to failure (\texttt{MN10S4,MN7S8}). The reason might lie in how RIDE encourages the agent to perform actions that affect the environment forcing the agent to assess all possible actions, so that the entropy in the policy distribution decays slowly. This hypothesis is buttressed by the results obtained in \texttt{MN7S4} and \texttt{MN10S4}: we recall that there are 3 useless actions in these scenarios (\texttt{pick up}, \texttt{drop} and \texttt{done}), and RIDE performs clearly worse (except for the $\_ep$ case in \texttt{MN7S4}). In complex scenarios, when those actions are relevant for the task, performance gaps between RIDE and ICM become narrower. 

\begin{figure}[h!]
    \centering
    \subfloat{\includegraphics[width=0.8\columnwidth]{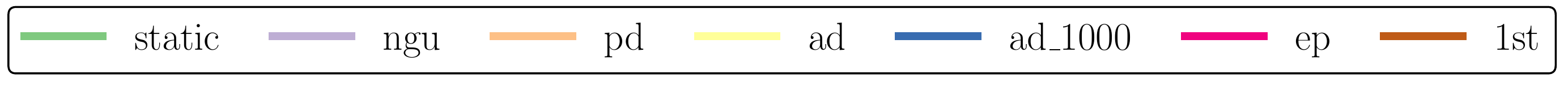}}\\ 
    \includegraphics[width=\columnwidth]{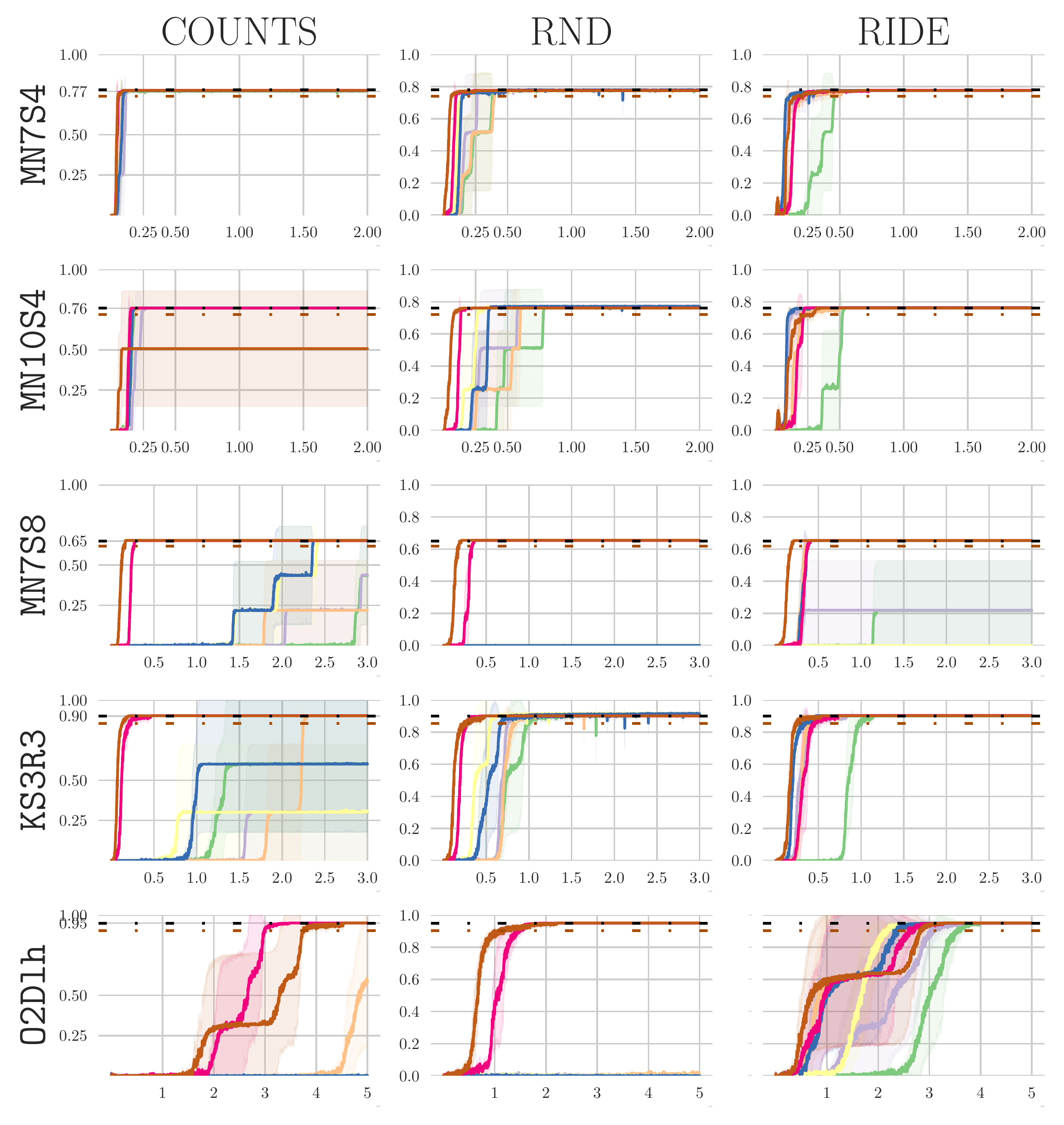}
    \caption{Convergence plots of the schemes reported in Tables \ref{table:rq1} and \ref{table:rq2}. Each column represents a Intrinsic Motivation type (COUNTS, RND and RIDE from left to right); each row represents the different scenarios (\texttt{MN7S4}, \texttt{MN10S4}, \texttt{MN7S8}, \texttt{KS3R3} and \texttt{O2Dlh}, from top to bottom). All figures depict the average extrinsic return as a function of the number of training steps/frames (in a scale of $1e7$). For each scenario, optimal and suboptimal scores are highlighted with horizontal black and brown lines, respectively.}
    \label{fig:preliminary_results}
\end{figure}

Finally, for the sake of completeness in the results exposed in RQ1 and RQ2, Figure \ref{fig:preliminary_results} shows the training convergence plots of COUNTS, RND and RIDE for different weighting and scaling strategies. These plotted curves permit to visually analyse the performance during the training process.

\subsection*{RQ3: Is the choice of the neural network architecture crucial for the agent's performance and learning efficiency?}

One of the most tedious parts when implementing an algorithm is to determine which network architectures to use. First of all, when using an actor-critic RL framework it is necessary to establish whether a single but two-headed network or two different (and independent) networks will be adopted for the actor and the critic modules. In addition, some IM approaches are based on neural networks to generate the intrinsic rewards.

Herein, we evaluate two of those solutions: RND and RIDE, evaluating the contribution of different neural network architectures to the overall performance of the agent. We use similar architectures to the ones used in RIDE \cite{raileanu2020ride} and RAPID \cite{zha2021rank}\footnote{Even with different neural architectures and base RL algorithms, they successfully solve the same tasks in MiniGrid with different sample-efficiency.}: (1) a two-headed shared actor-critic network built upon convolutional and dense layers and (2) two independent MLP networks for the actor and the critic, respectively (Figure \ref{fig:archs}). Moreover, we fix the RL algorithm (PPO) and detail the number of parameters and time taken for the forward and backward passes in each network for an informed comparison.

\begin{table}[h!]
    \centering
    \vspace{-3mm}
    \caption{Comparison between the network architectures described in Section \ref{subsec:architecures}.} 
    \resizebox{0.8\columnwidth}{!}
    {
     \begin{tabular}
     {L{3cm}R{2.2cm}R{2.2cm}R{2.2cm}R{2.2cm}}
     \toprule
     
     & \multicolumn{2}{c}{Lightweight (\emph{lw})} & \multicolumn{2}{c}{Default}\\
     \midrule
     & \makecell[rb]{Parameters} & \makecell[rb]{Time (ms)} & \makecell[rb]{Parameters} & \makecell[rb]{Time (ms)} \\
     \midrule
     \textit{Actor} & 14,087 & - & - &\\
     \textit{Critic} & 13,697 & - & - & \\
     \textit{Actor+Critic} & 27,784 & - &29,896 & -\\
     \midrule
     \textit{Dictionary} & - & 83.66 & - & 95.11 \\
     \textbf{Total COUNTS} & 27,784 & 724.25 & 29,896 & 937.37\\
     \toprule
     
     \textit{Embedding} & 13,632 & - & 19,392 & - \\
     RND & 27,264 & 336.39 & 38,784 & 721.64 \\
     \midrule
     \textbf{Total RND} & 55,048 & 986.13 & 68,937 & 1,408.42 \\
     \toprule
     
     \textit{Inverse} & 12,871 & - & 18,439 & -\\
     \textit{Forward} & 12,928 & - & 18,464 & -\\
     \textit{Embedding} & 13,632 & - & 19,392 & -\\
     RIDE & 39,431 & 388.84 & 56,295 & 844.43 \\
     \midrule
     \textbf{Total RIDE} & 67,215 & 1,177.75 & 86,191 & 1,791.70 \\
     \bottomrule
    \end{tabular}}
 \vspace{-3mm}
 \label{table:rq3_num_param}
\end{table}

First of all, Table \ref{table:rq3_num_param} informs about these details of the neural architectures in use for COUNTS, RND and RIDE. The table reports the differences in terms of the number of parameters of each network, and the latency taken by the sum of both forward and backward passes through those IM modules (we note that COUNTS uses a dictionary and not a neural network). In addition, we summarize the total number of parameters depending on the IM module that has been implemented, together with the actor-critic parameters. Referred to the total elapsed time, we report the total amount of time required for a rollout collection. This elapsed time takes into account both the forward and backward passes in the IM modules, and just the forward pass across the actor-critic, among other operations executed when collecting samples. Times are calculated when executing the experiments over an Intel(R) Xeon(R) CPU E3-1505M v6 processor running at 3.00GHz. 

\begin{table}[h]
    \centering
    \vspace{-5mm}
    \caption{Performance obtained with Counts, RND and RIDE when 1) using the default network configurations, 2) a lightweight architecture for the IM modules and keeping actor-critic with a default configuration ($\_lw\_im$), and 3) when both the IM and the actor-critic modules are implemented with the lightweight networks ($\_lw\_tot$). Values in the cells represent the training steps/frames (in a scale of 1e6) when the optimal average extrinsic return is achieved. Within brackets, the training steps when a suboptimal behavior is accomplished.} 
    \resizebox{\columnwidth}{!}
    {\begin{tabular}{L{2.5cm}C{2cm}C{2cm}C{2.2cm}C{2.2cm}C{2.2cm}}
     \toprule
     & \makecell[cb]{\texttt{MN7S4}} & \makecell[cb]{\texttt{MN10S4}} & \makecell[cb]{\texttt{MN7S8}} & \makecell[cb]{\texttt{KS3R3}} & \makecell[cb]{\texttt{O2Dlh}} \\
     \midrule
     COUNTS & 0.93 (0.86) & 1.87 (1.78) & $>$ 30 & $>$ 30 & $>$ 50 \\
     COUNTS$\_lw\_im$ & 0.93 (0.86) & 1.87 (1.78) & $>$ 30 & $>$ 30 & $>$ 50 \\
     COUNTS$\_lw\_tot$ & 1.64 (1.48) & 2.52 (2.36) & $>$ 30 (29.96) & $>$ 30 & $>$ 50 \\
     \midrule
     RND & 3.86 (3.79) & 7.84 (7.79) & $>$ 30 & 10.84 (9.72) & $>$ 50 \\
     RND$\_lw\_im$ & 5.66 (5.44) & 6.68 (6.61) & $>$ 30 & 10.97 (9.45) & $>$ 50 \\
     RND$\_lw\_tot$ & $>$ 20 & $>$ 20 & $>$ 30 & $>$ 30 & $>$ 50 \\
     \midrule
     RIDE & 2.49 (1.82) & 2.27 (2.14) & 4.01 (3.38) & 6.63 (4.39) & 30.88 (25.87) \\
     RIDE$\_lw\_im$ & 1.63 (1.31) & 1.75 (1.53) & $>$ 30 & 9.44 (5.08) & $>$ 50 \\
     RIDE$\_lw\_tot$ & 1.42 (1.05) & $>$ 20 & $>$ 30 & 8.00 (5.69) & $>$ 50 \\
     \bottomrule
    \end{tabular}}
    \vspace{-3mm}
 \label{table:rq3}
\end{table}

On the other hand, Table \ref{table:rq3} shows the performance of the agent when configured with such different network configurations. It can be seen that when reducing the number of parameters in both the actor-critic and the IM modules ($\_lw\_tot$), the agent's behaviour is critically deteriorated. This occurs even with COUNTS (Table \ref{table:rq3}), where the modification should have had less impact as the generation of intrinsic rewards does not depend on a neural network, but on a dictionary. When inspecting the performance of RIDE, its performance gets worse in all cases except for \texttt{MN7S4}, where the exploration requirements are the lowest among all the analyzed scenarios. Consequently, this modification can be less harmful. As for RND, the lightweight configuration of the networks makes the tasks not solvable by the agent.

Back again to the information shown in Table \ref{table:rq3_num_param}, it can be inferred that the number of parameters to be learned is mostly dependent on the IM networks under consideration, whereas joining the actor and the critic into a single two-headed network barely increases the dimensionality requirements\footnote{We note that the number of parameters is slightly increased, but they also differ in the type of layers that are used in each network (the two-headed network uses CNNs while the independent actor-critic only uses dense layers.}. Nevertheless, the time required to perform a forward pass increases in approximately 25\% when an unique actor-critic network is employed. Moreover, by using a single network, part of the parameters of the network are shared between the actor and the critic, which can induce more instabilities but also a faster learning (as the model may share features between the actor and the critic and require less samples to learn a given task). With this in mind, we carry out an additional ablation study considering only the reduction of parameters at IM modules, and maintaining the actor-critic as a single two-head network. 

Such results are provided in the 2nd row of every group of results in Table \ref{table:rq3} ($\_lw\_im$). These outcomes evince that when using RND$\_lw\_im$, slightly worse results are achieved with respect to RND with the default network setup. However, its performance does not degrade dramatically down to failure as with RND$\_lw\_tot$. Hence, using parameter sharing in a single network yields a faster learning process for this case. Regarding RIDE$\_lw\_im$, in some cases (\texttt{MN7S4} and \texttt{MN10S4}) it attains better results, whereas in \texttt{MN7S8} and \texttt{KS3R3} it suffers from a notorious performance decay (\texttt{MN7S8} is not solved). It can also be observed that the use of the single actor-critic network might be beneficial when reducing the complexity of the IM network ($\_lw\_im$), as it mitigates the performance degradation in 3 out of 5 scenarios (yet \texttt{MN7S8} and \texttt{O2Dlh} are not solved) when compared to separated actor-critic networks ($\_lw\_tot$), which fail over \texttt{MN7S8}, \texttt{O2Dlh} and \texttt{MN10S4}). 

\begin{figure}[h!]
    \centering
    \vspace{-3mm}
    \subfloat{\includegraphics[width=0.33\columnwidth]{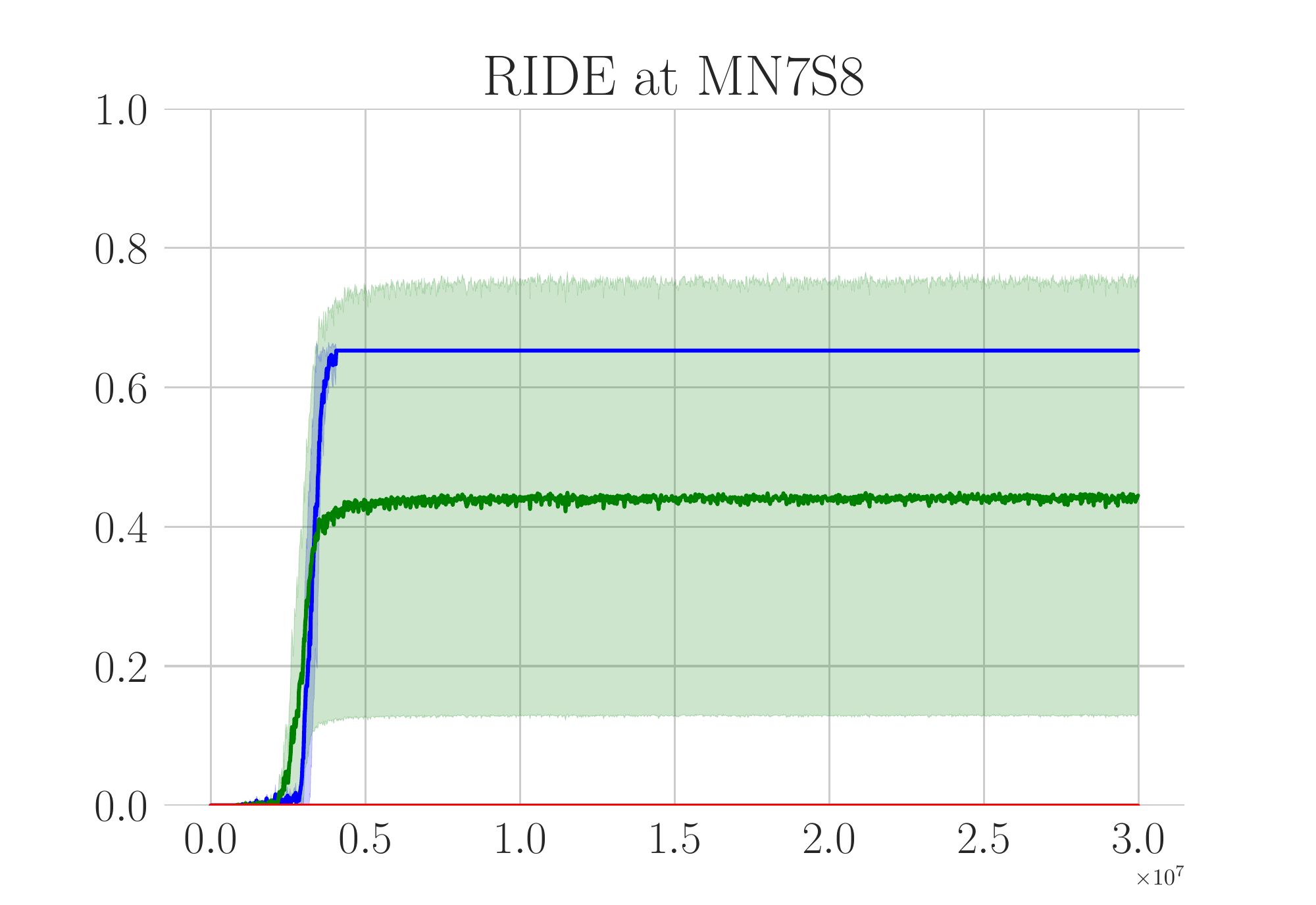}}
    \subfloat{\includegraphics[width=0.33\columnwidth]{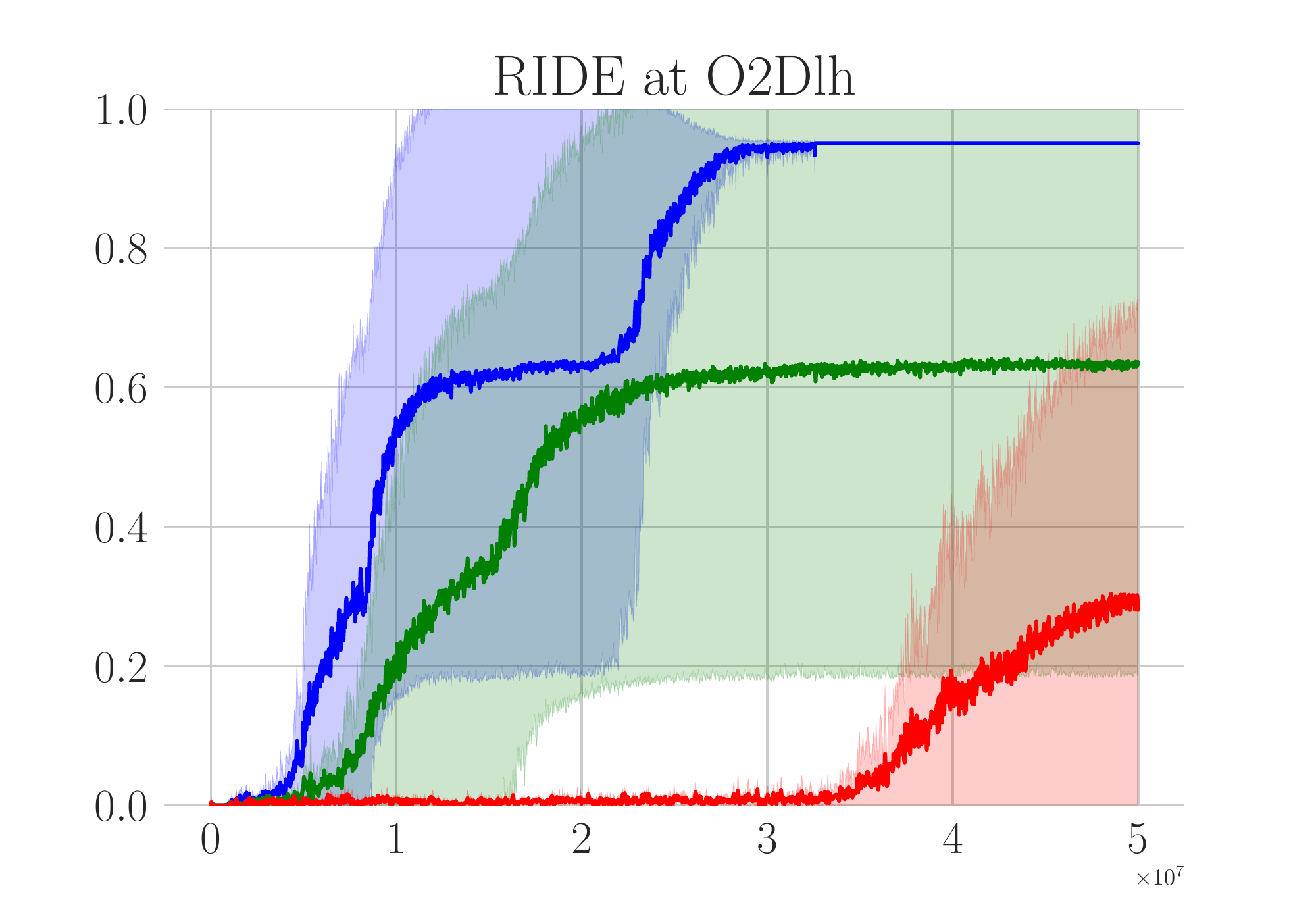}}
    \subfloat{\includegraphics[width=0.33\columnwidth]{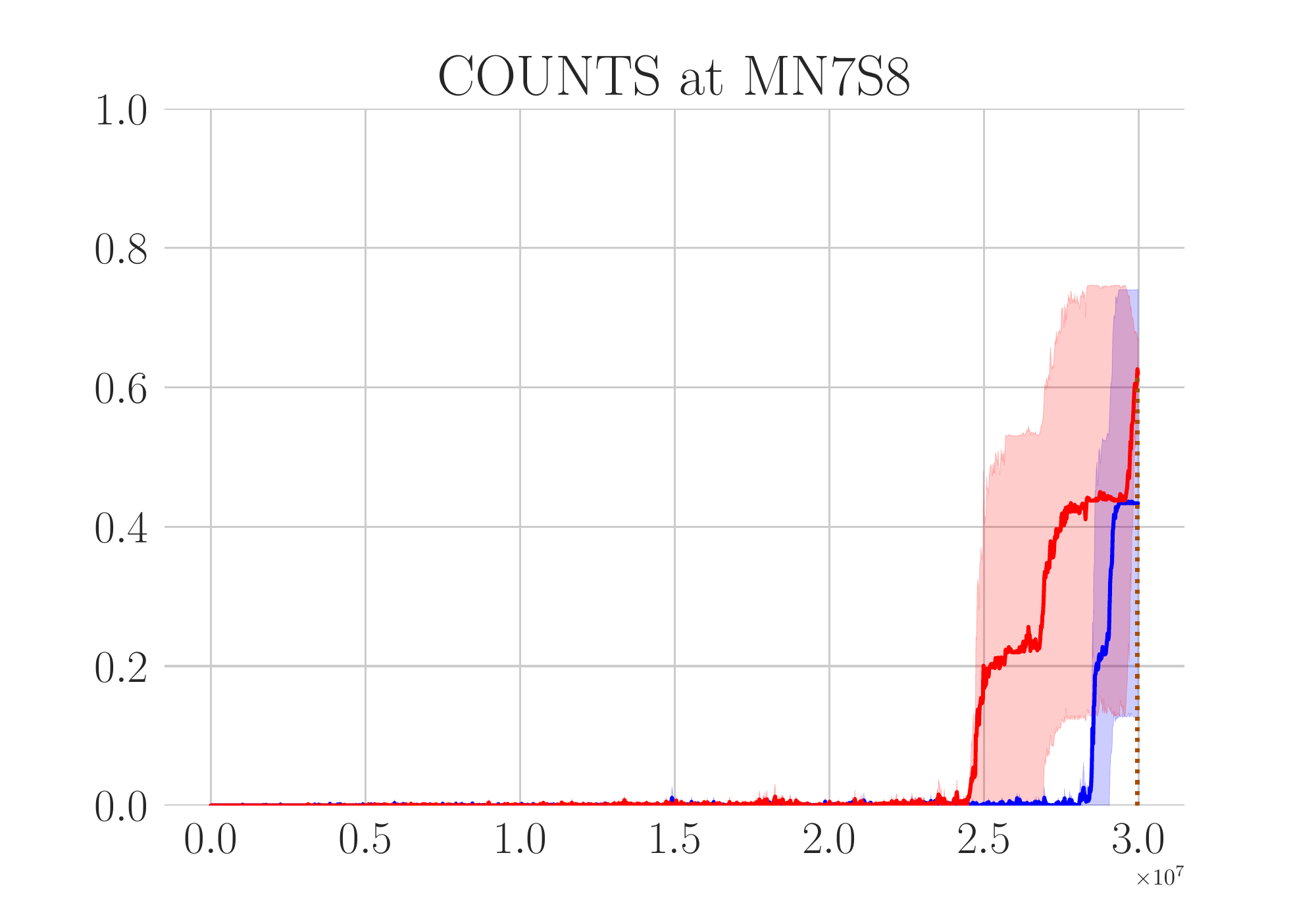}}
    \caption{Convergence plots of COUNTS and RIDE for some scenarios when using the default network (blue), $\_lw\_im$(green) and $\_lw\_tot$(red). All the figures depict the average extrinsic return as a function of the number of training frames.}
    \label{fig:rq3_plots}
\end{figure}

Finally, we include Figure \ref{fig:rq3_plots} in order to help the reader draw deeper conclusions and gain insights about the behaviour of the learning process. It can be seen that in the two cases in which RIDE$\_lw\_im$ failed (namely, \texttt{MN7S8} and \texttt{O2Dlh}), in two out of the three experiments that were run (\emph{seeds}) the agent learned to solve the task, which underscores the impact of using different actor-critic architectures. Moreover, with the default actor-critic approach and using the COUNTS approach, the agent is also able to solve the \texttt{MN7S8} task in two out of the three runs. When using COUNTS$\_lw\_tot$, the agent reaches suboptimal performance and almost the optimal one within the frame budget.

\vspace{5mm}
\section{Conclusion} \label{sec:conclusion}
In this work we have studied the actual impact of selecting different design choices when implementing IM solutions. More concretely, we have evaluated multiple weighting strategies to give different importance when combining the intrinsic and extrinsic rewards. Moreover, we have analysed the effect of applying distinct exploration degree levels
along with the influence of the complexity of the network architectures on the performance of both actor-critic and IM modules. To conduct the study we have utilized environments belonging to MiniGrid as benchmark to test the quality of proposed schemes in a variety of tasks demanding from hard to very hard intensity of exploratory behaviour.

\vspace{1mm}
On one hand, we have shown that using a static intrinsic coefficient might not be the best strategy if we focus on sample-efficiency. Adaptive decay strategies have proved to be the most promising ones, although they require a good parameterisation of the sliding window. 
Parameter decay approach, in turn, have performed competently but it is subject to a decay parametrisation which could be more dependent of the task at hand than the previous scheme, which makes this strategy more sensitive to the environment and the task to be solved (as it happens with $\epsilon-$greedy strategies in value-based methods). The use of multiple agents as in NGU \cite{badia2020never}, each featuring a different exploration-exploitation balance also suffers from this parametrisation but reports worse results.

\vspace{1mm}
On the other hand, the use of \textit{episode-level} exploration along with \textit{experiment-level} strategies seem to be preferable when having environments with hard exploration requirements. It is not a clear winner/preference between episodic counts and first visitation strategies as their performance is not only subject to the environment, but also to the selected IM strategy, although both achieve significant improvement in the performance. Hence, we encourage the implementation of any of these strategies in follow-up IM-related studies.

\vspace{1mm}
Last but not least, we have analyzed the impact of modifying the neural network architecture in both the actor-critic and IM modules. The results show that reducing the number of parameters at the IM modules deteriorate the performance of the agent, making it fail in some challenging scenarios which are feasible for the complex neural configuration. What is more, when reducing the IM network dimensions, it is preferable to use a shared two-headed actor-critic as it provides better results, although it is not clear whether those results are due to the use of a single neural network (and the underlying parameter sharing and common feature space for the actor and the critic) or to the adoption of different architectures (e.g. CNNs). Further research is necessary in this direction.

\vspace{1mm}
We hope this work can guide readers in the implementation of intrinsic motivation strategies to address tasks with (1) a lack of dense reward functions or (2) at hard exploration scenarios where the classic exploration techniques are insufficient. Aligned with the purposes of academic and industry communities, we make all the experiments available and provide the code to ensure reproducibility \cite{holzinger2019introduction}. In the future, we will intent to extend these analysis to more environments and algorithms in order to have more representative results.

\section*{Acknowledgments}
A. Andres and J. Del Ser would like to thank the Basque Government for its funding support through the research group MATHMODE (T1294-19) and the BIKAINTEK PhD support program. 

%
%


\bibliographystyle{unsrt} 
\bibliography{mybibliography}

\end{document}